\documentclass[letterpaper, 10 pt, conference]{ieeeconf}  

\IEEEoverridecommandlockouts                              

\usepackage{amssymb}
\usepackage{subcaption}

\usepackage{amsmath}
\usepackage{paralist}
\usepackage{tikz}
\usetikzlibrary{arrows,automata}
\usepackage{acronym}
\usepackage{amsmath}
\usepackage{amsthm}
\usepackage{amssymb}
\usepackage{mathtools}
\usepackage{amsfonts}
\usepackage{flushend}
 
\newif\ifuseboldmathops
\newif\ifuseittextabbrevs
\useboldmathopstrue   

\ifuseittextabbrevs

	\newcommand{\ie}{{\it i.e.}}

\else

	\newcommand{\ie}{i.e.~}

\fi

\ifuseboldmathops

\else

\fi

\ifuseboldmathops

\else

\fi

\ifuseboldmathops

\else

\fi

\ifuseboldmathops

\else

\fi

\newcommand{\calL}{\mathcal{L}}
\newcommand{\calF}{\mathcal{F}}
\newcommand{\calAP}{\mathcal{AP}}

\newcommand{\indicator}{\mathbf{1}}

\renewcommand{\vec}[1]{\mathbf{#1}}

\newcommand{\nat}{\mathbb{Z}}

\acrodef{mdp}[MDP]{Markov Decision Process}
\acrodef{pomdp}[POMDP]{Partially Observable Markov Decision Process}
\acrodef{ltl}[LTL]{Linear Temporal Logic}
\acrodef{dfa}[DFA]{Deterministic Finite Automaton}

\theoremstyle{definition}
 \newtheorem{definition}{Definition}
 \newtheorem{example}{Example}
\newtheorem{problem}{Problem}

\newtheorem{remark}{Remark}

\newcommand{\calA}{\mathcal{A}}

\acrodef{smdp}[Semi-MDP]{Semi-Markov decision process}
\acrodef{mcts}[MCTS]{Monte Carlo tree search}
\acrodef{uct}[UCT]{Upper Confidence Bound 1 applied to trees}
\acrodef{scltl}[scLTL]{syntactically co-safe LTL}
\acrodef{ssp}[SSP]{Stochastic Shortest Path}
\acrodef{p2sg}[SG(2)]{Two-player Stochastic Game}
\acrodef{mc}[MC]{Markov chain}
\acrodef{prefltl}[TPL]{ Temporal Preference Logic}
\acrodef{tld}[TLwD]{Temporal Logic with Distributions}
\acrodef{mtl}[Metric TL]{Metric Temporal Logic}
\acrodef{sta}[STA]{Stochastic Timed Automaton}

\newcommand{\dist}{\mathcal{D}}

\renewcommand{\Pr}{\mathbf{Pr}}

\newcommand{\prefmodel}{\mathfrak{M}}
\newcommand{\calM}{\mathcal{M}}

\newcommand{\APref}{\mbox{APF}}
\newcommand{\lex}{\mathsf{lex}}

 \newcommand{\lang}{\mathcal{L}}

\acrodef{gpf}[GPF]{generalized preference formula}

\acrodef{cp}[CP]{ceteris paribus}
\acrodef{milp}[MILP]{Mixed-Integer Linear Programming}
\acrodef{dfa}[DFA]{Deterministic Finite Automaton}

\title{\LARGE \bf   Probabilistic Planning  with Preferences over Temporal Goals}
\author{Jie Fu $^{1}$
\thanks{$^{1}$ Jie Fu is with the Department of Robotics Engineering, Worcester Polytechnic Institute, 01520, Worcester, USA. 
        {\tt\small jfu2@wpi.edu}}
    \thanks{This material is based upon work supported by the Air Force Office of Scientific Research under award number FA9550-21-1-0085.}
}

\begin{document}

\maketitle
\thispagestyle{empty}
\pagestyle{empty}

\begin{abstract}
We present a  formal language for specifying qualitative preferences over temporal goals and a preference-based planning method in stochastic systems.
Using automata-theoretic modeling, the proposed  specification allows us to express preferences over different sets of outcomes, where each outcome  describes a set of temporal sequences of  subgoals. We define the value of preference satisfaction given a stochastic process over possible outcomes and develop an algorithm for time-constrained probabilistic planning in labeled Markov decision processes where  an agent aims to   maximally satisfy its preference formula within a pre-defined finite time duration. We  present experimental results using a stochastic gridworld example and discuss  possible extensions of the proposed preference model.
\end{abstract}

\section{INTRODUCTION}
In an uncertain  environment, 
preference plays a key role in human's decision-making that decides what  constraints
to satisfy when not all constraints can be achieved
\cite{hastie2010rational}.  To achieve such cognitive flexibility in autonomous systems, it is necessary to investigate the model of preferences and the  planning problems given preferences in deterministic, probabilistic, or even adversarial environments.
In this paper, we focus on the specification of preferences over temporal goals, and the design of preference-based policies in a discrete stochastic system modeled as a \ac{mdp}.

The study of preference-based planning is motivated by existing probabilistic planning with temporal logic constraints. In these planning problems, the desired behavior of the system is first specified by a temporal logic formula \cite{manna2012temporal}, and then the agent  synthesizes a policy that either maximizing the probability of satisfying the formula \cite{ding2011mdp,hasanbeig2019reinforcement}, or incorporating the formula as constraints for probabilistic planning \cite{Lacerda2014,wen2015correct}, or treating the satisfaction of the formula as one objective in multi-criteria planning \cite{fuSynthesisSharedAutonomy2016}. However, a major limitation is that if the specification is not satisfiable or satisfied with a small probability, then a new specification will be generated and the process iterates. The lack of flexibility in specifying ``what needs to be satisfied'' motivates the development of the minimum violation planning concept: Starting with a user-defined preference relation to prioritized properties, minimum violation   planning in a deterministic system \cite{tumova2013least} decides which low-priority constraints to be violated. A related topic to minimal violation planning is called shield synthesis \cite{bloem2015shield,alshiekh2017safe}, where a shield performs run-time policy revision to ensure that the system satisfies a small set of critical properties, such as state-dependent safety conditions. Automated specification-revision is proposed in \cite{Lahijanian2016} where the formula can be revised with a cost and the planning problem is formulated into a multi-objective \ac{mdp} that trades off minimizing the cost of revision and maximizing the probability of satisfying the revised formula. 
Robust and recovery specifications are introduced by\cite{bloem2019synthesizing} and pre-specify what behaviors are expected when the part of the system specification (\ie,  the environment assumption) fails to be satisfied.   
 Minimal-violation planning, revision, and shield synthesis are all informed with  preferences.  In current work, a preference is either represented by a utility
 function \cite{hindriks2008using}, determined by prioritizing sub-specifications \cite{tumova2013least,Tomita2017},  a distance measure for the deviation from the original formula \cite{Lahijanian2016}, or a set of hard constraints for run-time verification. Yet, utilitarian representations and prioritization cannot generalize to general ordinal preferences, which can describe  pre-orders over possible outcomes.  In AI community,   preference-based
planning is studied and aims  to solve the most preferred plan as a sequence of actions
(see a survey \cite{baierPlanningPreferences2008}).   In
\cite{bienvenuSpecifyingComputingPreferred2011a}, the authors proposed
a logical language for specifying preferences over the evolution of
states and actions associated with a deterministic plan. 
 
 Motivated by existing work, we develop a model of preferences over temporal goals. This  model allows the agent to express its preference as \emph{a pre-order of sets of temporal evolving trajectories}, rather than a total order.  Based on the proposed model, we solve a class of time-constrained preference-based planning, in which the agent must maximally satisfy its preference  within a pre-defined finite time duration. The planning method is based on encoding the satisfaction of  preference formulas into the objective function and constraints of a \ac{milp}. By solving the MILP, we obtain a  finite-horizon policy that maximizes the preference satisfaction (defined formally in Sec.~\ref{sec:semantics}). We use a running example to explain the proposed preference specification languages and discuss the limitation in its expressiveness. Finally, we employ a stochastic gridworld example to show the outcomes of the preference-based planning methods given different preference specifications.
 
\section{Preliminary}
 \label{sec:prelim}
 In this section, we define the notations used throughout the paper and the model of the stochastic system. 
 
Notation: Given a finite set $X$, let $\dist(X)$ be the set of probability distributions over $X$.  Let $\Sigma$ be an alphabet (a finite set of symbols). Given $k \in \nat_{+}$, $\Sigma^k$ indicates a set of words with length $k$, $\Sigma^{\leq k}$ indicates a set of words with length no greater than $k$, and $\Sigma^0 = \lambda$ is the empty word. $\Sigma^\ast$ is the set of finite words (also known as the Kleene closure of $\Sigma$),  and $\Sigma^\omega$ is the set of infinite words. Given a set $X$, let $\indicator_X$ be the indicator function with $\indicator_X(x) =1, \text{ if } x \in X$ and $0$ otherwise.

 We consider the stochastic system is modeled by a labeled \ac{mdp} \cite{baier2008principles}. 
	A labeled \ac{mdp} is a tuple $M = \langle S, A, \nu, P, \calAP, L \rangle$, where $S$ and $A$ are finite state and action sets, $\nu\in \dist(S)$ is the initial state distribution, the transition probability function $P(\cdot \mid s, a) \in \dist(S)$ is defined as a probability distribution over the next state given action $a$ is taken at  state $s$, $\calAP$ denotes a finite set of atomic propositions, and $L: S \rightarrow 2^{\calAP}$ is a labeling function mapping each state to the set of atomic propositions true in that state.
  We defer the definition of policies to Section~\ref{sec:planning}. 

\section{Specifying Preferences over Temporal Goals}

\label{sec:semantics}
In this section, we introduce an automata-theoretic model of preferences over temporal goals. 
\begin{definition}[Preferences  of temporal goals]
The model of preferences of temporal goals is represented by a finite-state, deterministic automaton with preferences over acceptance conditions, formally, 
\[
\calA = \langle Q, 2^\calAP, \delta, q_0, \prefmodel \rangle,\]
where
\begin{itemize}
    \item $Q$ is a set of states. 
    \item $2^\calAP =\Sigma$ is the alphabet.
    \item $\delta: Q\times \Sigma \rightarrow Q$ is a deterministic transition function, which maps a state $q$ and an input symbol $\sigma \in \Sigma$ to a unique next state. 
    \item $q_0$ is the initial state.
    \item $\prefmodel=\langle \APref, \varphi \rangle$ is a preference model, 
where   
\begin{itemize}
\item $\APref$ is a set of atomic preference formulas. A preference formula is atomic if it can be expressed as the following form $X_0\preccurlyeq X_1 \preccurlyeq \ldots  \preccurlyeq X_n$ for some $X_i\subseteq Q$ and $X_i \cap X_j =\emptyset$, for any $i\ne j$, $0\le i, j\le 1$.   The formula is said to have a length $n$, which is the number of operators $\preccurlyeq$  in the formula.
The relation $\preccurlyeq$ is reflexive and transitive   (a so-called “preorder”) over $2^Q$ and its strict
sub-relation $\prec$ is given by:
\[
X \prec X' \mbox{iff } X \preccurlyeq X' \land X' \not \prec X.
\]
$\preccurlyeq$ is said to be total iff for all $X, X'$, either $X \preccurlyeq X'$ or $X' \preccurlyeq X$. 
\item $\varphi$ is a \ac{gpf} and is constructed from $\APref$ using propositional logic. A \ac{gpf} $\varphi$ is an atomic preference formula or one of the following two forms
\begin{enumerate}[(1)]
    \item $\psi_1 \& \psi_2\ldots\& \psi_n $ [General And].
    \item $\psi_1 | \psi_2\ldots | \psi_n $ [General Or].
\end{enumerate}
where $n\ge 1$ and each $\psi_i$ is a \ac{gpf}.
\end{itemize}
 \end{itemize}
 \end{definition}
 The ``General And'' ($\&$) and ``General Or'' ($\mid$) follow from \cite{bienvenuSpecifyingComputingPreferred2011a} and are different from logical $\land $ and $\lor$ that can be used to define atomic preference. Because $X_1\preccurlyeq X_2 \land X_2\preccurlyeq X_3$ is equivalently an APF $X_1\preccurlyeq X_2 \preccurlyeq X_3$ for the transitivity of  the preference relation. But $ \varphi_1 \&  \varphi_2$  means the combination of two different 
 preferences. This will be clarified later as we introduce the value of preference satisfaction (see Def.~\ref{def:sat_prob}).
 
\begin{remark}
Note that the negation of an atomic preference is not defined, because the negation of ``$X$ is preferred to $X'$'' does not mean that ``$X'$ is preferred to $X$''.  This is because if $X$ is not preferred to $X'$, then it can be one of the three possible cases:  \begin{inparaenum}[1)]\item $X'$ is preferred to $X$; \item $X$ and $X'$ are indifferent, or \item $X$ and $X'$ are incomparable.  \end{inparaenum}
\end{remark}

%
%

The   preference model $\calA$ can be used to specify preferences over subsets of words in $\Sigma^\ast$ where $\Sigma= 2^\calAP$. To see this,  we first consider a regular \ac{dfa}, that is,  $A = \langle Q, \Sigma, \delta, q_0,X\rangle$ with a reachability acceptance condition: a  word $w\in \Sigma^\ast$ is accepted by $A$ if $\delta(q_0, w)\in X$. 
Now, given  
  two sets $X\subseteq Q$ and $X'\subseteq Q$ and the states and transitions of the automaton $\calA$, we can derive two sets of words: $\lang(X)$ is the set of words accepted by $ \langle Q,\Sigma, \delta,q_0, X\rangle $  
and $\lang(X')$ is the set of words accepted by $ \langle Q,\Sigma, \delta,q_0, X'\rangle$. 
 The atomic preference $X\preccurlyeq X'$ means a word in $\lang(X')$ is preferred to any words in $\lang(X)$.  The interpretation of
 APF $X_0\preccurlyeq X_1 \preccurlyeq \ldots  \preccurlyeq X_n$ means a word in $\lang(X_i) $ is preferred to any words in $\lang(X_j)$ for $0 \le j < i$, for $1\le i \le n$.

%
%

We now use a running example to illustrate the proposed preference model.
\begin{example}
	\label{ex}
	Consider a robot motion planning example, the robot needs to sequentially visit a set of regions, denoted $A, B$ and $C$. 
	The preference of the human supervisor is stated informally as follows:
	\begin{itemize}
		\item P-A: The more regions being visited the better. 
		\item P-B: If only two regions are visited, he prefers visiting region $\{A,B\}$ to visiting either regions $\{B,C\}$, or $\{A,C\}$.
		\item P-C: If all three regions can be visited, he prefers to visit $A,B$ first and then $C$.   And visiting three regions (regardless of ordering) is better than visiting two regions.
	\end{itemize}
	
	We first construct the automaton in Fig.~\ref{fig:dfa-ex}. The automaton is understood as follows: State $0$ means no regions has been reached. State $1,2,3$ means that the agent has visited $A,B,C$, respectively. State $4,5,6$ means that the agent has visited two of the regions. State $7$  means that the agent has visited all three regions. The atomic preferences are 
	\begin{itemize}
		\item ($P_{0,X}$) $X \succ \emptyset$ for all $X\subseteq Q\setminus \{q_0\}$;
		\item ($P_1$) $\{7\} \succcurlyeq \{4,6\}$;
		\item ($P_2$) $\{7\}\succcurlyeq\{5\}$;
		    \item ($P_3$) $\{5\}\succcurlyeq \{4,6\} $;
		        \item ($P_4$) $\{4,6\}\succcurlyeq \{1,2,3\}$;
		\item ($P_5$) $ \{7\} \succcurlyeq \{4,6,5\} \succcurlyeq \{1,2,3\}\succcurlyeq \{0\} $;
	\end{itemize}
It is clear that P-A is expressed by the APF $P_5$ and P-B is expressed by APF $P_3$.
	However, P-C cannot be defined using the automaton in Fig.~\ref{fig:dfa-ex} because the set of states in automaton cannot distinguish the set of words that visiting $\{A,B\}$ and then $C$. To express the last preference, we  refine the automaton by re-partitioning the Myhill–Nerode  equivalence relation \cite{automatabook2006} over $\Sigma^\ast$ defined by the structure of the automaton in Fig.~\ref{fig:dfa-ex} to generate a new automaton in Fig.~\ref{fig:dfa-ex-refine}. 
	With the new automaton, we  add a new atomic preference $P_6\coloneqq \{8\}\succcurlyeq \{7\}$ and duplicate the original $P_2$ as $P_2' \coloneqq \{8\}\succcurlyeq \{5\}$. The last preference  is then  expressed as
	\begin{itemize}
		\item P-C: $  P_1 \& P_2'\&  P_6$.
	\end{itemize}

	\begin{figure}[!ht]
		\centering
		\vspace{-4ex}
		\begin{tikzpicture}[->,>=stealth',shorten >=1pt,auto,node distance=2.5cm,    el/.style = {inner sep=2pt, align=left, sloped},
        scale = 0.5,transform shape]

  \node[state,initial] (0) {$0$};
  \node[state] (1) [above right of=0] {$1$};
  \node[state] (2) [right of=0] {$2$};
  \node[state] (3) [below right of=0] {$3$};
  \node[state] (4) [above right of=2] {$4$};
  \node[state] (5) [right of=2] {$5$};
  \node[state] (6) [ right of=3] {$6$};
  \node[state] (7) [right of=5] {$7$};

  \path (0) edge              node {$A$} (1)
        (0) edge              node {$B$} (2)
        (0) edge              node {$C$} (3)
        (1) edge              node {$C$} (4)
        (1) edge              node {$B$} (5)
        (2) edge              node {$A$} (5)
        (2) edge              node {$C$} (6)
        (3) edge  [bend left=130]            node  [above,pos=0.9] {$A$} (4)
        (3) edge              node {$B$} (6)
        (4) edge              node {$B$} (7)
        (5) edge              node {$C$} (7)
        (6) edge              node {$A$} (7);

\end{tikzpicture}
 
		\caption{The automaton for defining preferences.}
				\vspace{-4ex}
		\label{fig:dfa-ex}
	\end{figure}
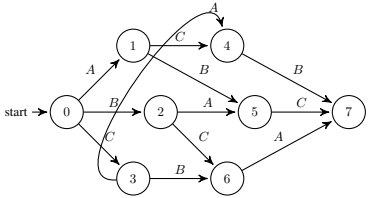
	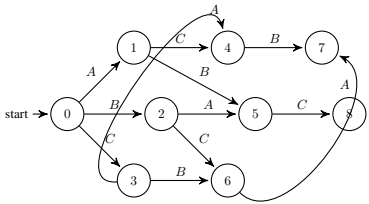
\begin{figure}[!ht]
		\centering
				\vspace{-4ex}
		\begin{tikzpicture}[->,>=stealth',shorten >=1pt,auto,node distance=2.5cm,
        scale = 0.5,transform shape]

  \node[state,initial] (0) {$0$};
  \node[state] (1) [above right of=0] {$1$};
  \node[state] (2) [right of=0] {$2$};
  \node[state] (3) [below right of=0] {$3$};
  \node[state] (4) [above right of=2] {$4$};
  \node[state] (5) [right of=2] {$5$};
  \node[state] (6) [ right of=3] {$6$};
  \node[state] (7) [right of=4] {$7$};
  \node[state] (8) [right of=5] {$8$};
  \path (0) edge              node {$A$} (1)
        (0) edge              node {$B$} (2)
        (0) edge              node {$C$} (3)
        (1) edge              node {$C$} (4)
        (1) edge              node {$B$} (5)
        (2) edge              node {$A$} (5)
        (2) edge              node {$C$} (6)
(3) edge  [bend left=130]       node  [above,pos=0.9]{$A$} (4)
(3) edge              node {$B$} (6)
        (4) edge              node {$B$} (7)
        (5) edge              node {$C$} (8)
        (6) edge      [bend right=100]         node  [left,pos=0.8]{$A$} (7);

\end{tikzpicture}
 
				\vspace{-2ex}
		\caption{The refined automaton for defining preferences.}
		\label{fig:dfa-ex-refine}
	\end{figure}
	
\end{example}




\begin{remark} In our formulation, we directly express the preferences over acceptance conditions in an automaton. This can be related to preferences over temporal logic formulas as follows:
Given the states and transitions of a \ac{dfa}, with different final state sets, the automaton accepts different   languages, \ie subsets of $\Sigma^\ast$. Each subset $\calL$ of $\Sigma^\ast$ can be associated with a temporal logic formula such that a word is in the set $\calL$ if and only if the word satisfies the said formula. Conversely, if we have two  formulas $\varphi_1$ and $\varphi_2$ such that the word satisfying $\varphi_1$ is preferred to any word satisfying $\varphi_2$, we can construct the union product of $\calA_1$ (expressing $\varphi_1$) and $\calA_2$ (expression $\varphi_2$), and then define the preferences over subsets of states in the union product. The  \ac{gpf} would enable us to specify preference,  such as, $\varphi_1$ is preferred to $\varphi_2$ and $\varphi_3$ is preferred to $\varphi_2$ but $\varphi_1$ and $\varphi_3$ are incomparable. Constructing automated algorithms to translate preferences over temporal logic formulas to our preference models of temporal goals is our ongoing work.
\end{remark}

In a stochastic system, each preference formula can be satisfied to some degree. We define the value of preference satisfaction  as follows.
\begin{definition}[Value of Preference Satisfaction]
\label{def:sat_prob}
Given a stochastic process $\{Y_t, t\ge 0 \}$  with a stopping time defined on the probability space $(\Sigma , \calF, P)$,  the value of preference satisfaction $v:\Phi \rightarrow [0,1]$ is defined as follows
\begin{itemize}
    \item For each atomic preference $p\in \APref$ and $p\coloneqq X_0 \preccurlyeq X_1\preccurlyeq X_2 \ldots \preccurlyeq X_n$,
  \begin{equation}
      \label{eq:definePref}
      v(p) = \left\{ \begin{array}{cc}
         \Pr( X_i) &  \text{ if } \exists i >0, \Pr(  X_i)\ge
         \Pr(  X_{i-1}) \\ 
         & \forall  k \ge i ,\Pr(  X_k) <
         \Pr(  X_{k-1})  \\
         0& \text{otherwise.} 
    \end{array}\right.
  \end{equation} 
  where $\Pr(Z) \triangleq \Pr(\{w\in \Sigma^\ast\mid \delta(q_0, w)\in Z\})$ is the probability of terminating in set $Z$ upon reading the word sampled from the stochastic process.
 \item For   GPF $\varphi_1\& \varphi_2$, $v(\varphi) = \min(v(\varphi_1) , v(\varphi_2) ).$
  \item For   GPF $\varphi_1\mid \varphi_2$, $v(\varphi) = \max(v(\varphi_1) , v(\varphi_2) ).$
\end{itemize}
\end{definition}
Here the stopping time is a random variable that determines when to stop the process.  With a stopping time, the sampled word is of a finite length.


Next, we provide some important remarks about the expressiveness of the proposed preference models:
\begin{itemize}
	\item Conditional preferences: A conditional preference means if some condition is satisfied, then the preference is expressed. For example, if the fuel is still sufficient after visiting two regions $A,B$, then the preference of visiting region $D$ over $C$ is expressed. If the fuel is insufficient after visiting $A,B$, then the agent is indifferent to $C$ and $D$ and wishes to visit one more region before running out of fuel.  To express such conditional preferences, we would need to introduce quantitative formulas and cost/reward functions, which is out of the scope for this paper. The current definition of preferences can capture qualitative conditional preferences. For example, if $A,B$ is visited, then $D$ is preferred to $C$. This preference can be expressed by first defining the set $L_1$ of words ending in $D$ after seeing only $A,B$, and a set $L_2$ of words ending in $C$ after seeing only $A,B$, and construct the \ac{dfa} that accepts $L_1\cup L_2$. We can define the preferences over state sets in the constructed \ac{dfa} to express the conditional preference. 
	\item The quality of satisfaction: Consider the stochastic process $\{Y_t,t\ge 0\}$ and two preference formulas $\varphi_1$ and $\varphi_2$. In some cases, it is desirable to say that if the value $v(\varphi_1)$ is below a given threshold, then the agent's preference is $\varphi_2$, otherwise, the agent's preference is $\varphi_1$. This preference cannot be expressed with the given model as it requires the generalization of the model with quantitative formulas.   
\end{itemize}

\section{Preference-based  Probabilistic Planning with Time Constraints}
\label{sec:planning}
Preference satisfaction is often coupled with limited resources (monetary or time). In this section, we consider a class of planning problems with preferences and time constraints.
First, to reason about system dynamics and preferences over temporal sequences, we use the product operation \cite{baier2008principles} to augment the state space of a labeled \ac{mdp} with automaton states.
  \begin{definition}[Product \ac{mdp}]
Given a labeled \ac{mdp} $M = \langle S,A,\nu ,P,\calAP, L\rangle$, a \ac{dfa} with preference $\calA = \langle Q, \Sigma, \delta, q_0, \prefmodel \rangle$, the product of $M$ and $\calA $ is denoted by \[
\calM=  M \otimes \calA  = \langle  S \times Q ,A, \bar \nu, \Delta  \rangle, \] with (1) a set $S\times Q$ of states, (2) the set $A$ of actions, (3) a distribution over the initial states $ \bar \nu(s_0, \delta(q_0,L(s_0)) = \nu(s_0)$, (3) the transition function defined by $ \Delta((s',q') \mid (s, q), a) = P(s' \mid s, a) \indicator_{\{q'\}}(\delta(q,L(s')))$.
\end{definition} 

 In the product \ac{mdp}, a finite-horizon, nonstationary, stochastic policy in the \ac{mdp} is a sequence $\pi = \mu_{0}\mu_1\ldots \mu_T $ where $\mu_t: S\times Q \rightarrow \dist(A)$ that maps a state $(s,q)$ into a distribution over  actions at time $t$, for some $0\le t\le T$. Given a product \ac{mdp} $\calM$ and a policy $\pi$, the policy induces a stochastic process $\calM^{\pi} = \{(S_t,Q_t) \mid t = 0, \cdots, T \}$ where $S_t$ and $Q_t$ are the random variables for the $t$-th MDP state and automaton states, respectively. It holds that $(S_{t+1} ,Q_{t+1})\sim \Delta(\cdot \mid (S_t,Q_t), A_t)$ and $A_t \sim \mu_t(S_t,Q_t)$. We say the stochastic process $\calM^\pi$ is policy $\pi$-induced. 

\begin{definition}[Finite-Horizon Preference Satisfaction]
Given a Markov chain $\calM^\pi$ and a \ac{gpf} $\varphi$, the value of preference satisfaction given a finite time bound $T$, denoted $v(\varphi, \pi,T) $,
is defined to be the value of $\varphi$ given the  stochastic process $\{L(S_t),0\le t< T\}$ where $S_t$ is the $t$-th \ac{mdp} state in $\calM^\pi$.
\end{definition}

The problem of preference-based planning with time constraint is formally stated as follows.

\begin{problem}
\label{prob:statement}
Given a labeled \ac{mdp} $M$, a preference model $\calA$, a \ac{gpf} $\varphi$, and a finite time $T$, compute a policy $\pi=\mu_0\mu_1\ldots \mu_{T-1}$ that maximizes the value $v_\varphi$ (see Def.~\ref{def:sat_prob}) for satisfying the preference formula within $T$ time steps. Formally, $
\pi^\ast = \arg\max_\pi v(\varphi, \pi,T).
$
\end{problem}

To solve Problem~\ref{prob:statement}, we combine time-constrained \ac{mdp} planning with the semantics of preferences to formulate a mixed integer linear program (MILP). However, it is noted that an atomic preference with length greater than $1$ will introduce a lexicographic constraint and requires iterative search. In the scope of this work, we consider a subclass of preference models where the atomic preferences are of length  1 and the solution vith MILP.

\noindent \textbf{Decision variables} We introduce a set of continuous variables $\{y(t, (s,q),a)\mid t =0,1, \ldots, T-1, (s,q)\in S\times Q, a\in A \}$ where $y(t,(s,q),a)$ is interpreted as the probability of visiting state $(s,q)$ in the process  $\calM^\pi$ and taking action $a$ at time $t$, for $t=0,\ldots, T$.  The variables satisfy the following two constraints:
\begin{align}
    & \sum_{a\in A}y({0},(s,q),a)  = \bar \nu(s,q),\quad \forall (s,q) \in S\times Q, \label{eq:initial} \\
   &  \sum_{a\in A}y(t ,(s',q'),a) =   \nonumber\\ 
   &\sum_{a\in A}\sum_{(s,q)\in S\times Q}\Delta((s',q')|(s,q),a)y(t-1,(s,q),a) \nonumber\\
  &\forall t: 1 \leq t \leq   T -1, \; \forall (s',q') \in S\times Q. \label{eq:balance}
\end{align}
where \eqref{eq:initial} enforces the  constraint from the initial state distribution and \eqref{eq:balance} means that the probability of reaching $(s',q')$ at time $t$ is the same as the sum of probabilities of reaching some other state at time $t-1$ from which an action  is chosen to reach $(s',q')$ next. These two constraints are generated based on the linear program formulation for constrained \ac{mdp}s \cite{altman1999constrained}.

\noindent \textbf{Encoding of Preferences}
Given a \ac{gpf} $\varphi$, we introduce a Boolean variable $z_\varphi$, which equals to $1$ if there exists a policy $\pi$ that $v(\varphi,\pi,T)>0$.
We recursively generate the \ac{milp} constraints corresponding
to $\varphi$. The time $T$ is omitted with the understanding that the following constraint generation is applicable for time-constrained planning.

\noindent \textbf{1) Atomic Preferences of Length 1} 
Each atomic preference $p \coloneqq (X\preccurlyeq X')$ of length 1 is associated with a value $v_p\in [0,1]$, subject to the following constraint:  
\begin{align}
&v_p  - y(T,X')  \ge M\cdot( z_p-1), \nonumber\\
&v_p - y(T,X')\le 0, v_p \ge 0, v_p\le M\cdot z_p,\nonumber \\
&y(T,X')-y(T,X)\le M\cdot z_p- \epsilon(1-z_p)\nonumber\\
&y(T,X') - y(T,X)\ge m(1-z_p) + \epsilon z_p \nonumber
\end{align}
where $y(T,X) =\sum_{q \in X} \sum_{a\in A} y(T,(s,q),a)$ is the probability of reaching the set $X$
 at time $T$, $M>0$ is the upper bound on $y(T,X')-y(T,X)$, which is one, and  $m$ is the lower bound on $y(T,X')-y(T,X)$, which is  $-1$, and $\epsilon >0$ is a constant close to zero. The last two constraint means if $y(T,X') \ge y(T,X)$ then $z_p=1$, else $z_p=0$. The first and second constraints mean that $v_p=y(T,X')$ when $z_p=1$. The third and fourth constraints mean that $v_p=0$ if $z_p=0$. The strong preference $X\prec X'$ is expressed using the same set of constraints. This is because when $z_p=1$, the last constraint enforces $y(T,X') > y(T,X)$ when $\epsilon > 0$.

\noindent \textbf{2) Boolean operation}
We now encode   $\& $ (general and) and  $\mid$ (general or) of atomic preferences as mixed-integer linear constraints. Let an atomic preference $p$ have its corresponding variable $v_p$.  The ``generalized and'' of atomic preferences $\varphi\coloneqq p_1\& p_2$ generates the following constraints:
\begin{align*} 
 & v_\varphi \le v_{p_i}, \quad i =1,2\\
 & v_\varphi -v_{p_1}\ge m \cdot (1-z), \\
 & v_\varphi -v_{p_2} \ge m \cdot z\\
 & v_{p_1}-v_{p_2} \le M\cdot z -\epsilon(1- z)
\end{align*}
where $z$ is a Boolean variable such that when if $v_{p_1} \ge v_{p_2}$ then $z=1$. The first three lines of constraints enforce $v_\varphi = \min(v_{p_1},v_{p_2})$.   
The ``generalized or'' of atomic preferences $\varphi\coloneqq p_1\mid p_2$ generates the following constraints,
\begin{align*} 
 & v_\varphi \ge v_{p_i}, \quad i =1,2\\
 & v_\varphi \le v_{p_1}+ M\cdot (1-z),\\
 & v_\varphi \le v_{p_2} + M\cdot z,\\
 & v_{p_1}-v_{p_2} \le M\cdot z -\epsilon(1- z),
 \end{align*}
where $z$ is a Boolean variable such that when if $v_{p_1} \ge v_{p_2}$ then $z=1$. The first three lines of constraints enforce $v_\varphi = \max(v_{p_1},v_{p_2})$.
For general conjunction/disjunction of more than one atomic preference formulas, we can apply the constraint generation recursively. 

Finally, for a given \ac{gpf} $\varphi$, we obtain a set of constraints $\{f_i(\vec{v}, \vec{y}, \vec{z})\ge 0, i=1,\ldots k\}$ where $\vec{z}$ is a vector of Boolean variables, $\vec{y}$ is the vector of decision variables, and $\vec{v}$ is the vector of values of sub-formulas in the \ac{gpf}.
The \ac{milp} formulation for time-constrained preference-based planning is shown as follows: 
\begin{align*}
    \max_{\vec{y},\vec{v},\vec{z}} & \quad  v_\varphi\\
    \text{s.t.: }&  \eqref{eq:initial}, \eqref{eq:balance}, f_i(\vec{v}, \vec{y}, \vec{z})\ge 0, i=1,\ldots k\\
& \vec{z} \in \{0,1\}^\ell,  \vec{v} \succeq 0, \vec{y} \succeq 0.
\end{align*} 
where $k, \ell$ are appropriate numbers generated from the set of constraints, and $\succeq $ is the inequality symbol for element-wise no less than.
The solution to the  \ac{milp}  defines the optimal policy  of Problem~\ref{prob:statement}: For each  $t$ such that $ 0 \leq t \leq   T -1$, the decision rule is 
\begin{multline}
\mu_{t}((s,q),a):=\frac{y(t,(s,q),a)}{\sum_{a'}y(t,(s,q),a')},\\
\forall\; (s,q) \in S\times Q \text{ and } \forall a \in A. 
\end{multline}
 
It is noted that for time-constrained planning, the optimal policy is time-dependent \cite{mundhenkComplexityFiniteHorizonMarkov2000}.
\begin{remark}
   In  \cite{bienvenuSpecifyingComputingPreferred2011a}, the authors introduces aggregated preferences. One such preference is  lexicographical preference, written as $\lex(\varphi_1,\ldots, \varphi_n)$ where each $\varphi_i$ is a  preference. This preference means there is a lexicographic ordering over preference formulas. Only if $\varphi_i$ is not satisfiable, then the agent should try to satisfy $\varphi_{i+1}$, for $i=1,\ldots,n-1$.  To generate policies given a lexicographical preferences, we can use multiple rounds of preference-based planning given  \ac{gpf}s $\varphi_1,\ldots,\varphi_n$, following the lexicogrphical ordering. The computation is terminated when the first satisfiable \ac{gpf} and its corresponding optimal policy are found. 
\end{remark}
\paragraph*{Complexity} The planning method  given the automata-encoding of preferences has the same complexity as \ac{milp}s, which are NP-hard.  
As the optimal policy is time-dependent, the planning state space is the Cartesian product of $S$, $Q$, and $\{t\mid T-1 \ge t\ge 0\}$. Thus, the computation can be expensive when the formula is complex and has a long time horizon. 

\section{Case Study}
Continuing with the preferences specified in Example~\ref{ex}, let's consider the planning problem in a stochastic grid world as shown in Fig.~\ref{fig:small_gridworld}. For agent’s different action (heading north (‘N’), south
(‘S’), west (‘W’) and east (‘E’)), the probability of arriving the intended cell is shown in Table~\ref{tab:dynamics}. 
In Fig.~\ref{fig:small_gridworld}, blue disk represents the agent whose initial cell is $10$, the green cells are regions of interests, with the labeling: $L(0)=A$, $L(12)=B$, and $L(22)=C$, and the red cells are obstacles. If the agent reaches an obstacle, then it will stay in the cell for all future time. We consider the same automaton in Fig.~\ref{fig:dfa-ex}, with preferences over a set of temporal goals. 

  \begin{minipage}{\textwidth}
\begin{minipage}[b]{0.25\textwidth}
    \centering
    \scalebox{0.9}{
    \begin{tabular}{c|cccc}
         Action & N &E&S&W  
        \\ \hline\\
        `N'& 0.8 &0.1&0 &0.1\\
        `E' & 0.05 &0.8&0.15&0\\
        `S' & 0&0.1 &0.7&0.2\\
         `W' &0.15& 0&0.15&0.7
    \end{tabular}}
    \captionof{table}{The probabilistic dynamics. }
    \label{tab:dynamics}
\end{minipage}
  \begin{minipage}[b]{0.22\textwidth}
    \centering
    \includegraphics[width=0.7\textwidth]{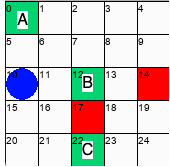}
    \captionof{figure}{A  $5\times 5$ gridworld.}
    \label{fig:small_gridworld}
\end{minipage} 
\end{minipage}

\begin{figure*}
    \centering
    \begin{subfigure}[b]{0.25\textwidth}
    \includegraphics[width= \textwidth]{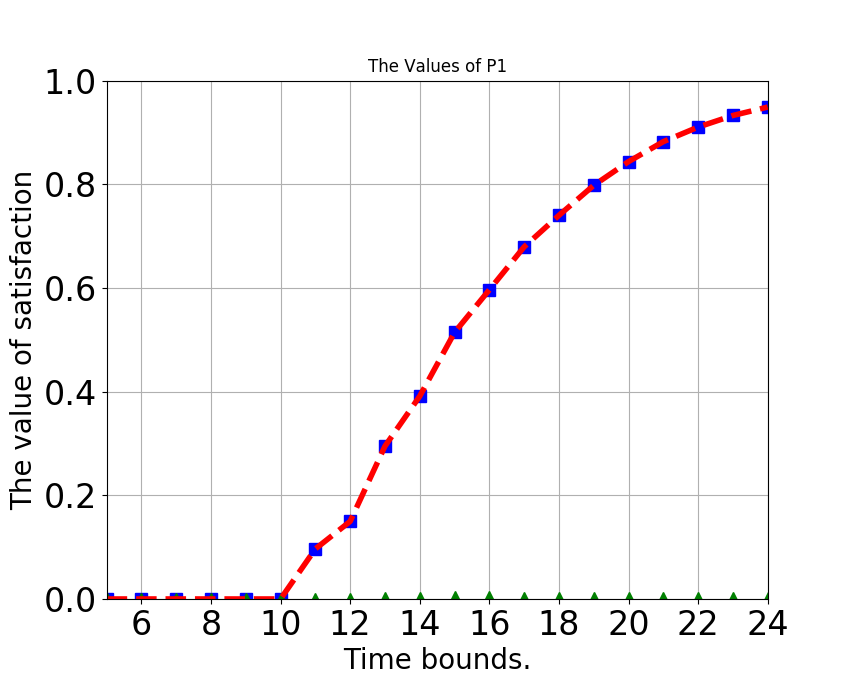}
    \caption{The value of satisfaction for $\varphi_1  $. }
    \label{fig:p1}
    \end{subfigure}
    \begin{subfigure}[b]{0.23\textwidth}
    \includegraphics[width=\textwidth]{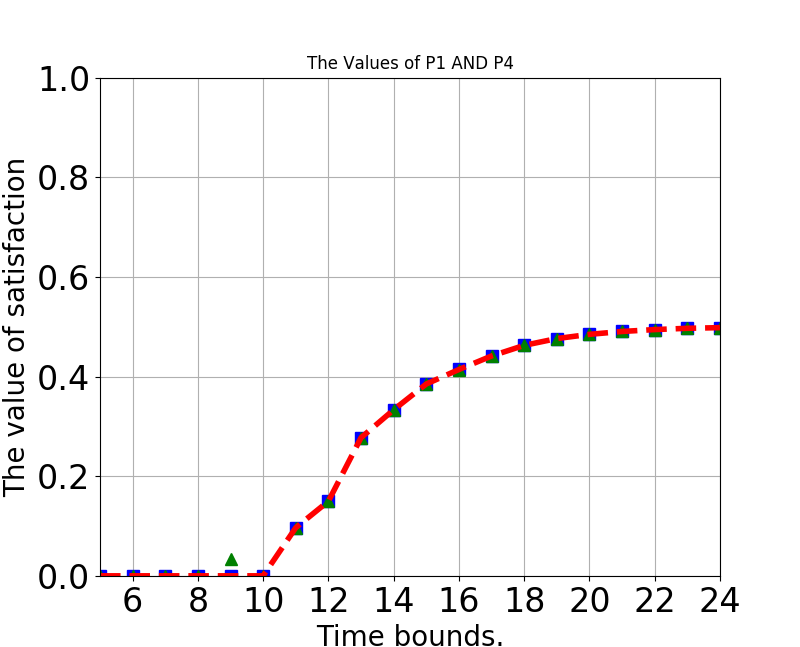}
    \caption{The value of satisfaction for $\varphi_2 = P_1\& P_4$.}
    \label{fig:p1ANDp4}
    \end{subfigure}
    \begin{subfigure}[b]{0.23\textwidth}
    \includegraphics[width=\textwidth]{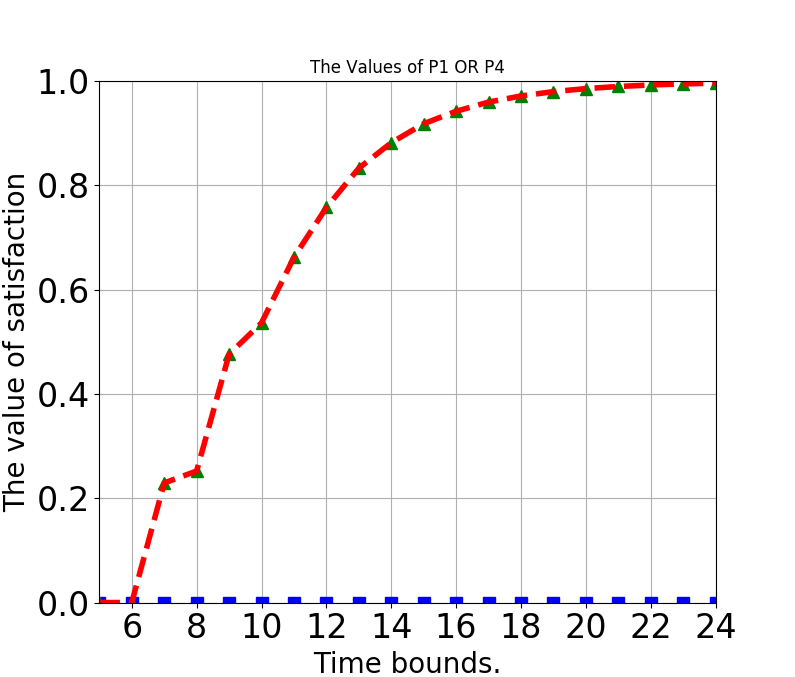}
    \caption{The value of satisfaction for $\varphi_3 = P_1\mid P_4$.}
    \label{fig:p1ORp4}
    \end{subfigure}
     \begin{subfigure}[b]{0.25\textwidth}
    \includegraphics[width=\textwidth]{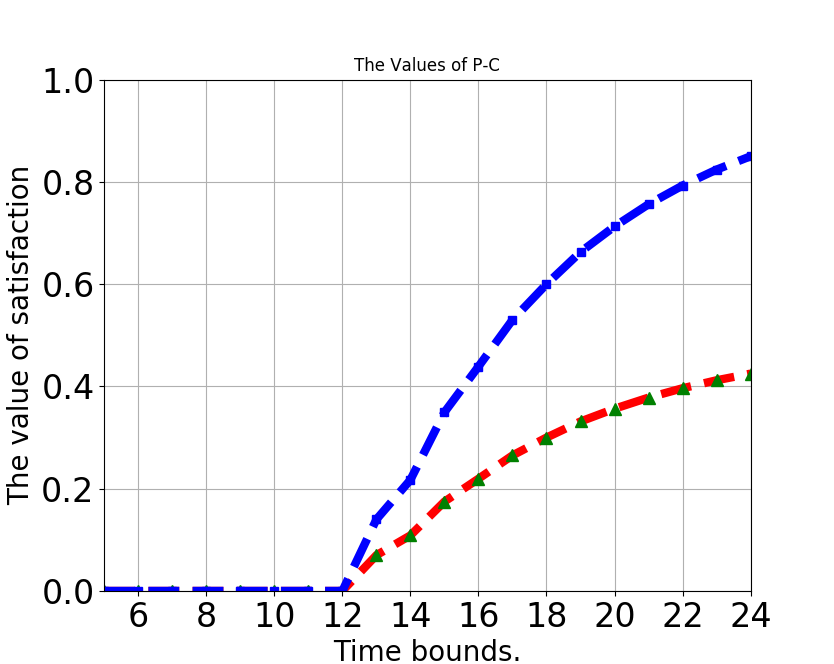}
    \caption{The value of satisfaction for P-C $: P_1\& P_2'\& P_6$.}
    \label{fig:pc}
    \end{subfigure}
    \caption{The values of satisfaction given different preference formulas and different time bounds.}
    \label{fig:value3}
\end{figure*}

We first consider three preference specifications: 1) $\varphi_1\triangleq P_1$, 2) $\varphi_2 \triangleq P_1\& P_4$, and 3) $ \varphi_3 \triangleq P_1\mid P_4$ (see the definition of preferences in Example~\ref{ex}). We vary the bound $T$ on the total time steps. Figure~\ref{fig:value3} shows the optimal value of $v(\varphi, \pi, T)$ under the optimal policy for $\varphi \in \{\varphi_1,\varphi_2,\varphi_3\}$.

In Fig.~\ref{fig:p1}, the green triangles (resp. blue squares) represent the probabilities of ending in $S\times \{4,6\}$ (resp. $S\times \{7\}$) given different time limits $T$ for $5\le T\le 24$ under the computed policies.  The red dashed line represents the values of satisfaction for $\varphi=P_1$ at different $T$. When $T$ is small $\le 10$, the agent cannot visit all three regions to reach the automaton state $7$, thus, the value of satisfaction is zero. When $T\ge 11$, the agent can satisfy the preference. The value of satisfaction increases monotonically as $T$ increases. There are some small probabilities ($\le 0.002$) of ending in $S\times \{4,6\}$ when $T\ge 11$.

In Fig.~\ref{fig:p1ANDp4} and Fig.~\ref{fig:p1ORp4}, the    green triangles (resp. blue squares) represent the  values of $P_1$ (resp. $P_4$) under the optimal policies. The red dashed line represents the values of satisfaction for $\varphi=P_1\& P_4$ (resp. $\varphi = P_1\mid P_4$) given different $T$. In the case of conjunction (Fig.~\ref{fig:p1ANDp4}), it is observed that  the value of satisfaction is lower than that for the case of disjunction (Fig.~\ref{fig:p1ORp4}). This is understandable, as to satisfy both $P_1$ and $P_4$, the agent is required to ensure, when the time stops, it will has a larger chance of visiting two regions than one, and at the same time, ensures the probability of visiting three regions is greater than that of visiting two.

Comparing Fig.~\ref{fig:p1ORp4} and Fig.~\ref{fig:p1}, it is noted that in the case of disjunction, the values of $P_1$ are all zeros (blue squares) at different time bounds $T$. This is because for any $T>0$, the value of satisfying $P_4$ is greater than the value of satisfying $P_1$ given the optimal policy. Due to the disjunction, the agent satisfies $\varphi_3$ by satisfying $P_4$ for $T\ge 6$.

 Next, we consider a more complex specification $\varphi \coloneqq \mbox{P-C}$ which says that the agent prefers to reach region $A,B$ first and then $C$ if all three regions can be visited. We use the automaton in Fig.~\ref{fig:dfa-ex-refine} which distinguishes a set of words that contains $A,B$ before seeing a $C$  from the rest of words containing all three symbols. The result is shown in Fig.~\ref{fig:pc}, the red, dashed line is the values of satisfying $\mbox{P-C}$ under different time bounds. The sequences of green triangles are the values of $P_1$ given the optimal policy, which coincide the values of $\mbox{P-C}$ given the same policy under different $T$. The blue dotted line is the sum $v(P_1)+v(P_2')$ under the optimal policy. The sum is the probability of visiting all three regions in the policy-induced stochastic process. Before time step $12$, the specification is not satisfiable and the values are zero. This is because the added constraint, $\{8\}\succcurlyeq  \{7\}$, is not satisfiable when $T\le 12$.
 

From the experiment, we can observe a major difference in preference-based planning
and probabilistic planning with temporal logic constraints \cite{baier2008principles,hasanbeig2019reinforcement,ding2011mdp}. If the goal is to simply reaching all three regions, then the optimal policy should be generated by optimally satisfying a preference formula $\{7\}\succcurlyeq \{0\}$ in \ac{dfa} (Fig.~\ref{fig:dfa-ex}) based on Def.~\ref{def:sat_prob}, 
 which would attain a value close to 1 as $T$ increases to infinity. With  preferences, other than a constraint that requires the probability of visiting three regions is greater than that of visiting none, we can incorporate more flexible constraints,  for example, visiting $k$ regions  is preferred to visiting $j$  when the time runs out for any $k>j$.

Experiments are conducted using Python MIP toolbox \cite{MipPythonTools} on a MacBook   with 1.6 GHz Intel Core i5 and 8 GB Memory. The maximal CPU time for running all experiments is $28$ sec (for $T=24$ and preference formula P-C).

\section{Conclusion and Future Work}

In this paper, we presented a way to specify preferences over temporal goals and a synthesis method to generate the optimal policy for the value of preference satisfaction
in a stochastic system. This preference model supports preferences over temporally extended goals and is built on the relation between automata theory and preferences in deterministic planners \cite{bienvenuSpecifyingComputingPreferred2011a}.  
Building on this work, we will further investigate the planning algorithms for more general atomic preferences (with length greater than one). We will study the specifications of temporally evolving preferences,  generalized conditional preferences, and their corresponding algorithms in  probabilistic   and stochastic systems.
\bibliographystyle{IEEEtran}
\bibliography{refs}

\begin{thebibliography}{10}
\providecommand{\url}[1]{#1}
\csname url@rmstyle\endcsname
\providecommand{\newblock}{\relax}
\providecommand{\bibinfo}[2]{#2}
\providecommand\BIBentrySTDinterwordspacing{\spaceskip=0pt\relax}
\providecommand\BIBentryALTinterwordstretchfactor{4}
\providecommand\BIBentryALTinterwordspacing{\spaceskip=\fontdimen2\font plus
\BIBentryALTinterwordstretchfactor\fontdimen3\font minus
  \fontdimen4\font\relax}
\providecommand\BIBforeignlanguage[2]{{%
\expandafter\ifx\csname l@#1\endcsname\relax
\typeout{** WARNING: IEEEtran.bst: No hyphenation pattern has been}%
\typeout{** loaded for the language `#1'. Using the pattern for}%
\typeout{** the default language instead.}%
\else
\language=\csname l@#1\endcsname
\fi
#2}}

\bibitem{hastie2010rational}
R.~Hastie and R.~M. Dawes, \emph{Rational choice in an uncertain world: The
  psychology of judgment and decision making}.\hskip 1em plus 0.5em minus
  0.4em\relax Sage, 2010.

\bibitem{manna2012temporal}
Z.~Manna and A.~Pnueli, \emph{The temporal logic of reactive and concurrent
  systems: Specification}.\hskip 1em plus 0.5em minus 0.4em\relax Springer
  Science \& Business Media, 2012.

\bibitem{ding2011mdp}
X.~C. Ding, S.~L. Smith, C.~Belta, and D.~Rus, ``Mdp optimal control under
  temporal logic constraints,'' in \emph{2011 50th IEEE Conference on Decision
  and Control and European Control Conference}.\hskip 1em plus 0.5em minus
  0.4em\relax IEEE, 2011, pp. 532--538.

\bibitem{hasanbeig2019reinforcement}
M.~Hasanbeig, Y.~Kantaros, A.~Abate, D.~Kroening, G.~J. Pappas, and I.~Lee,
  ``Reinforcement learning for temporal logic control synthesis with
  probabilistic satisfaction guarantees,'' in \emph{2019 IEEE 58th Conference
  on Decision and Control (CDC)}.\hskip 1em plus 0.5em minus 0.4em\relax IEEE,
  2019, pp. 5338--5343.

\bibitem{Lacerda2014}
B.~{Lacerda}, D.~{Parker}, and N.~{Hawes}, ``Optimal and dynamic planning for
  markov decision processes with co-safe ltl specifications,'' in \emph{2014
  IEEE/RSJ International Conference on Intelligent Robots and Systems}, 2014,
  pp. 1511--1516.

\bibitem{wen2015correct}
M.~Wen, R.~Ehlers, and U.~Topcu, ``Correct-by-synthesis reinforcement learning
  with temporal logic constraints,'' in \emph{2015 IEEE/RSJ International
  Conference on Intelligent Robots and Systems (IROS)}.\hskip 1em plus 0.5em
  minus 0.4em\relax IEEE, 2015, pp. 4983--4990.

\bibitem{fuSynthesisSharedAutonomy2016}
J.~Fu and U.~Topcu, ``Synthesis of {{Shared Autonomy Policies With Temporal
  Logic Specifications}},'' vol.~13, no.~1, pp. 7--17, Jan. 2016.

\bibitem{tumova2013least}
J.~Tumova, G.~C. Hall, S.~Karaman, E.~Frazzoli, and D.~Rus, ``Least-violating
  control strategy synthesis with safety rules,'' in \emph{Proceedings of the
  16th international conference on Hybrid systems: computation and
  control}.\hskip 1em plus 0.5em minus 0.4em\relax ACM, 2013, pp. 1--10.

\bibitem{bloem2015shield}
R.~Bloem, B.~K{\"o}nighofer, R.~K{\"o}nighofer, and C.~Wang, ``Shield
  synthesis,'' in \emph{International Conference on Tools and Algorithms for
  the Construction and Analysis of Systems}.\hskip 1em plus 0.5em minus
  0.4em\relax Springer, 2015, pp. 533--548.

\bibitem{alshiekh2017safe}
M.~Alshiekh, R.~Bloem, R.~Ehlers, B.~K{\"o}nighofer, S.~Niekum, and U.~Topcu,
  ``Safe reinforcement learning via shielding,'' \emph{The Thirty-Second AAAI
  Conference on Artificial Intelligence}, 2018.

\bibitem{Lahijanian2016}
M.~Lahijanian and M.~Kwiatkowska, ``Specification revision for {{Markov}}
  decision processes with optimal trade-off,'' in \emph{Proc. 55th Conference
  on Decision and Control ({{CDC}}'16)}, 2016, pp. 7411--7418.

\bibitem{bloem2019synthesizing}
R.~Bloem, H.~Chockler, M.~Ebrahimi, and O.~Strichman, ``Synthesizing reactive
  systems using robustness and recovery specifications,'' in \emph{2019 Formal
  Methods in Computer Aided Design (FMCAD)}.\hskip 1em plus 0.5em minus
  0.4em\relax IEEE, 2019, pp. 147--151.

\bibitem{hindriks2008using}
K.~V. Hindriks and M.~B. Van~Riemsdijk, ``Using temporal logic to integrate
  goals and qualitative preferences into agent programming,'' in
  \emph{International Workshop on Declarative Agent Languages and
  Technologies}.\hskip 1em plus 0.5em minus 0.4em\relax Springer, 2008, pp.
  215--232.

\bibitem{Tomita2017}
T.~Tomita, A.~Ueno, M.~Shimakawa, S.~Hagihara, and N.~Yonezaki, ``{Safraless
  LTL synthesis considering maximal realizability},'' \emph{Acta Informatica},
  vol.~54, pp. 655--692, 2017.

\bibitem{baierPlanningPreferences2008}
\BIBentryALTinterwordspacing
J.~A. Baier and S.~A. McIlraith, ``Planning with {{Preferences}},'' \emph{AI
  Magazine}, vol.~29, no.~4, p.~25, 2008. [Online]. Available:
  \url{https://www.aaai.org/ojs/index.php/aimagazine/article/view/2204}
\BIBentrySTDinterwordspacing

\bibitem{bienvenuSpecifyingComputingPreferred2011a}
M.~Bienvenu, C.~Fritz, and S.~A. McIlraith,
  ``\BIBforeignlanguage{en}{Specifying and computing preferred plans},''
  \emph{\BIBforeignlanguage{en}{Artificial Intelligence}}, vol. 175, no. 7-8,
  pp. 1308--1345, May 2011.

\bibitem{baier2008principles}
C.~Baier and J.-P. Katoen, \emph{Principles of model checking}.\hskip 1em plus
  0.5em minus 0.4em\relax MIT press, 2008.

\bibitem{automatabook2006}
J.~E. Hopcroft, R.~Motwani, and J.~D. Ullman, \emph{Introduction to Automata
  Theory, Languages, and Computation (3rd Edition)}.\hskip 1em plus 0.5em minus
  0.4em\relax USA: Addison-Wesley Longman Publishing Co., Inc., 2006.

\bibitem{altman1999constrained}
E.~Altman, \emph{Constrained Markov decision processes}.\hskip 1em plus 0.5em
  minus 0.4em\relax CRC Press, 1999, vol.~7.

\bibitem{mundhenkComplexityFiniteHorizonMarkov2000}
M.~Mundhenk, J.~Goldsmith, C.~Lusena, E.~Allender, N.~M. Mundhenk, and N.~J.
  Goldsmith, \emph{Complexity of {{Finite}}-{{Horizon Markov Decision Process
  Problems}}}, 2000.

\bibitem{MipPythonTools}
S.~T.A.M, H. G. {and}~Toffolo, ``Mip: {{Python}} tools for {{Modeling}} and
  {{Solving Mixed}}-{{Integer Linear}} {{Programs}} ({{MIPs}}),'' 2020.

\end{thebibliography}
\end{document}